\documentclass{article}

\usepackage{arxiv}
\usepackage[utf8]{inputenc} 
\usepackage[T1]{fontenc}    
\usepackage[hidelinks]{hyperref}       
\usepackage{url}            
\usepackage{booktabs}       
\usepackage{amsfonts}       
\usepackage{nicefrac}       
\usepackage{microtype}      
\usepackage{graphicx}
\usepackage{doi}
\usepackage{acro}           
\acsetup{make-links}        
\usepackage{amsmath}        
\usepackage{nicematrix}
\usepackage{abraces}

\DeclareAcronym{ai}{
  short=AI,
  long=Artificial Intelligence,
}
\DeclareAcronym{bom}{
  short=BOM,
  long=Bill of Materials,
}
\DeclareAcronym{cop}{
  short=COP,
  long=Combinatorial Optimization Problem,
}
\DeclareAcronym{dqn}{
  short=DQN,
  long=Deep Q-Networks,
}
\DeclareAcronym{drl}{
  short=DRL,
  long=Deep Reinforcement Learning,
}
\DeclareAcronym{erp}{
  short=ERP,
  long=Enterprise Resource Planning,
}
\DeclareAcronym{fjsp}{
  short=FJSP,
  long=Flexible Job Shop Scheduling Problem,
}
\DeclareAcronym{gnn}{
  short=GNN,
  long=Graph Neural Network,
}
\DeclareAcronym{jssp}{
  short=JSSP,
  long=Job Shop Scheduling Program,
}
\DeclareAcronym{mdp}{
  short=MDP,
  long=Markov Decision Process,
}
\DeclareAcronym{mes}{
  short=MES,
  long=Manufacturing Execution System,
}
\DeclareAcronym{pdr}{
  short=PDR,
  long=Priority Dispatching Rule,
}
\DeclareAcronym{ppo}{
  short=PPO,
  long=Proximal Policy optimization,
}
\DeclareAcronym{rl}{
  short=RL,
  long=Reinforcement Learning,
}

\title{Optimizing Job Shop Scheduling in the Furniture Industry: A Reinforcement Learning Approach Considering Machine Setup, Batch Variability, and Intralogistics}

\author{Malte Schneevogt,\,M.Sc.\\
        (Lead Author)\\
        Faculty of Wood Technology and Construction\\
        Rosenheim Technical University of Applied Sciences\\
        Hochschulstr. 1, 83024 Rosenheim\\
        \texttt{malte.schneevogt@stud.th-rosenheim.de}\\
	 \And
	 {Dipl.-Ing\,(FH)\,Karsten Binninger,\,M.Sc.}\\
      (Supervising Scientist)\\
	 Center for Research, Development and Transfer\\
	 Rosenheim Technical University of Applied Sciences\\
	 Hochschulstr. 1, 83024 Rosenheim\\
	 \texttt{karsten.binninger@th-rosenheim.de}\\
	 \And
	 {Prof.\,Dr.-Ing.\,Noah Klarmann}\\
      (Supervising Professor)\\
	 Faculty of Management and Engineering\\
	 Rosenheim Technical University of Applied Sciences\\
	 Hochschulstr. 1, 83024 Rosenheim\\
	 \texttt{noah.klarmann@th-rosenheim.de}\\
}

\renewcommand{\shorttitle}{}

\hypersetup{
pdftitle={Optimizing Job Shop Scheduling in the Furniture Industry: A Reinforcement Learning Approach Considering Machine Setup, Batch Variability, and Intralogistics},
pdfsubject={eess.SY, cs.LG},
pdfauthor={Malte Schneevogt},
pdfkeywords={Job Shop Scheduling, Production Scheduling, Reinforcement Learning, Markov Decision Process },
}

\begin{document}

\maketitle

\begin{abstract}
This paper explores the potential application of Deep Reinforcement Learning (DRL) in the furniture industry. In order to offer a broad product portfolio, most furniture manufacturers are organized as a job shop, which ultimately results in the Job Shop Scheduling Problem (JSSP). The JSSP is addressed with a focus on extending traditional models to better represent the complexities of real-world production environments. Existing approaches to JSSPs frequently fail to consider critical factors such as machine setup times, varying batch sizes, intralogistics, buffer capacities, or deadlines, which are essential in industrial settings. In order to overcome these limitations, a concept for a model is proposed that incorporates these elements, providing a higher level of information detail to enhance scheduling accuracy and efficiency.
The concept introduces the integration of DRL for production planning, which is particularly suited to batch production industries such as the furniture industry. The model extends traditional approaches to JSSPs by including job volumes, buffer management, transportation times, and machine setup times. This enables more precise forecasting and analysis of production flows and processes, accommodating the variability and complexity inherent in real-world manufacturing processes.
In a training environment the Reinforcement Learning (RL) agent learns to optimize scheduling decisions. The agent operates within a discrete action space, making decisions based on detailed observations of machine states, job volumes, and buffer statuses. A reward function, specifically tailored to the specific industrial applications, guides the agent's decision-making process, thereby promoting efficient scheduling and meeting production deadlines.
Two integration strategies for implementing the RL agent are discussed: episodic planning, which is suitable for low-automation environments, and continuous planning, which is ideal for highly automated plants. While episodic planning can be employed as a standalone solution, the continuous planning approach necessitates the integration of the agent with Enterprise Resource Planning (ERP) and Manufacturing Execution Systems (MES). This integration enables real-time adjustments to production schedules based on dynamic changes.
\end{abstract}


\section{Introduction}

The optimization of production planning is a crucial aspect of industrial manufacturing processes. It increases the overall production efficiency, ensures timely delivery, and reduces costs by utilizing resources effectively. In recent years, \ac{rl} has emerged as a powerful tool in solving complex optimization problems across various domains. Its ability to learn optimal policies through interaction with environments has revolutionised diverse domains, showcasing its versatility and effectiveness. In robotics, \ac{rl} has been utilized for autonomous navigation, manipulation, and control tasks \cite{Aydin.2000}\cite{Lillicrap.2015}, in gaming, it demonstrates superhuman performance in complex games, such as Go \cite{Silver.2016} or Dota 2 \cite{OpenAI.}. \ac{rl} offers several advantages over traditional optimization methods, including its adaptability to dynamic environments, its ability to handle large state and action spaces, and its capability to learn from experience without requiring explicit problem knowledge. These advantages make it an ideal tool for industrial process optimization. 

Due to its high level of complexity, the furniture industry requires a particularly diligent and optimized production planning. Its products are composed of numerous individual components, each of which undergoes separate manufacturing processes. A furniture article can be made from hundreds of components, each with an individual path through the production. Some components are simply purchased parts that do not require any further processing in the factory, while other parts involve complex manufacturing processes among many different machines. These processes need to be coordinated and integrated during production, particularly if various components or articles are produced on the same production lines, resulting in the batch production of products or components. In order to produce a wide range of products, most furniture shops are designed as job shops. This setup offers a high degree of flexibility for batch production, but inherently leads to the \ac{jssp}, a \ac{cop} known from process optimization.

In a \ac{jssp} a set of $n$ jobs $J$ = \{$J_{0}$, $J_{1}$, $J_{2}$, $...$, $J_{n}$\} is to be processed on $m$ machines $M = \{M_1, M_2, ..., M_m\}$. Each machine can only process one operation at a time. Each job is assigned with a specific machine sequence that must be followed during the production of the particular product. For each machine, the operations have specific processing times $d_{ij}$ where $i \in (1, m)$ and $j \in (1, n)$. The total number of operations $O$ is $n \times m$. The number of possible schedules in a \ac{jssp} is growing exponentially by $(n!)^m$. Common scheduling heuristics such as First-Come-First-Serve or Earliest-Due-Date-First become ineffective as the number of jobs and machines increases, and computation times become unfeasible due to the exponential growth of combinatorial possibilities. Algorithms such as Genetic Algorithms \cite{Pezzella.2008}, Simulated Annealing \cite{vanLaarhoven.1992}, or Taboo Search \cite{Taillard.1994} struggle to effectively solve the \ac{jssp} due to its NP-hard nature and the presence of complex dependencies among operations and resources \cite{Pinedo.2012}.

\begin{figure}[ht]
    \centering
    \includegraphics[width=10cm]{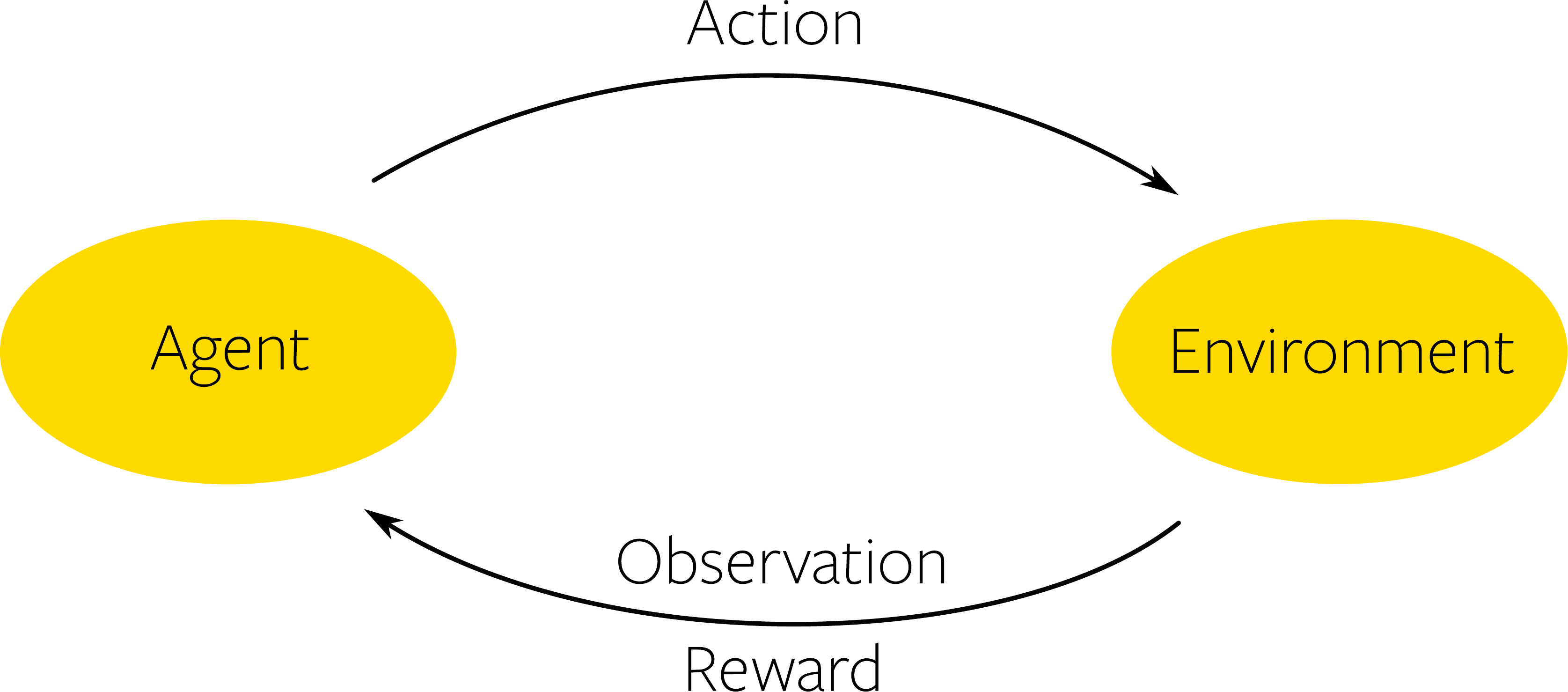}
    \caption{Reinforcement Learning Circuit \cite{Sutton.2018}}
    \label{fig:RLcircuit}
\end{figure}

\ac{drl} was successfully applied to the \ac{jssp} on several occasions, finding competitive solutions for \ac{jssp} benchmark problems \cite{Waschneck.2018,Han.2020,Liu.2020,Zhang.2020,Wang.2021,SerranoRuiz.2024}. \ac{drl} is a machine learning technique that combines deep learning and \ac{rl} to enable agents to learn and make decisions in complex environments by using neural networks to process and act on high-dimensional data. The agent takes actions based on the observations and its policy, receives rewards as feedback, and adjusts its neural network parameters in order to maximize cumulated future rewards, the so-called return (cf. Figure \ref{fig:RLcircuit}). This process enables the agent to learn optimal behaviours for complex tasks in high-dimensional spaces.

Advances in deep learning techniques, particularly the development of deep neural networks, have significantly improved the ability of \ac{rl} algorithms to handle high-dimensional and continuous state spaces. Due to their ability to approximate complex value functions and policies, \ac{drl} algorithms, such as \ac{dqn} and \ac{ppo}, have demonstrated remarkable performance in challenging tasks. The availability of large-scale computing resources, coupled with distributed computing frameworks such as TensorFlow or PyTorch, has enabled researchers to efficiently train \ac{rl} agents even on complex environments. This scalability has facilitated the exploration of \ac{rl} in real-world applications, where complex decision-making processes are prevalent.
The \ac{jssp} is a sequential decision-making process, and thus it can be modelled as a \ac{mdp}. This mathematical framework is commonly used in \ac{ai} and operations research for decision-making in uncertain environments where probabilistic state transitions can be influenced by the agent's actions.
While a generic \ac{jssp} provides a good theoretical model for scheduling in general, real-world production environments often have additional complexities that are not captured in the JSSP model.
This paper presents a framework for a \ac{drl} model that addresses the complexity of a real-world production environment and demonstrates how the proposed model can be applied in the furniture industry.

\section{State of the Art} \label{cha:SOA}

As early as 1995, Zhang und Dietterich \cite{Zhang.1995} explored the application of \ac{rl} techniques to the \ac{jssp} by further developing the results of the scheduling algorithms by Deale et al. \cite{Deale.1994}, who used a simulated annealing approach for job shop scheduling. The system learns to make informed decisions that lead to more efficient schedules and thereby shows the potential of \ac{rl} to solve \ac{jssp}s. 

In 2000, Aydin and Öztemel \cite{Aydin.2000} presented an approach that is composed of a simulated job shop environment and a \ac{rl} agent that selects the most appropriate priority rule from a set of available rules to assign jobs to machines. The aim of their work is to provide a flexible and adaptive approach to job shop scheduling, capable of handling a dynamic manufacturing environment, coming one step closer to a fully automated, intelligent manufacturing system.

Gabel and Riedmiller \cite{Gabel.2012} introduced a novel approach to solve \ac{jssp}s in 2012, by using distributed policy search \ac{rl}. This multi-agent approach treats the \ac{jssp} as a series of sequential decision-making tasks, where each \ac{rl}-agent operates autonomously and uses a probabilistic dispatching policy for decision making. These dispatching policies are represented by a small set of real-valued parameters that are continuously adapted and refined, to enhance the overall performance of the scheduling process. Although the computation time was reduced, the achieved solutions were not better than those of conventional solvers.

In 2008, Pezzella et al. \cite{Pezzella.2008} presented a genetic algorithm that solves the \ac{fjsp}. Unlike in a \ac{jssp}, the operations of a \ac{fjsp} can be assigned to one of several machines, instead of only one specific machine. Due to the additional decision layer of selecting machines for each operation, the complexity of a \ac{fjsp} is even higher than in a \ac{jssp}. The presented algorithm outperformed existing models and traditional dispatching rules. Their approach utilizes \ac{drl} to tackle the problem more effectively by including innovative approaches for representation learning and policy learning. They demonstrated that genetic algorithms are effective in solving \ac{fjsp}s.

Foundational research in the field of \ac{drl} for continuous control tasks was presented by Lillicrap et al. \cite{Lillicrap.2015} in 2015. This research expands the capabilities of deep learning beyond the discrete domain to tasks where actions can take any value within a continuous range. They employ an actor-critic architecture where the actor generates actions, and the critic evaluates them based on a learned value function. Also, they use a replay buffer to store and reuse past experiences to mimic successful techniques.

Shahrabi et al. \cite{Shahrabi.2017} address the complex problem of dynamic job shop scheduling, where job arrivals are unpredictable and machine breakdowns occur randomly. The paper presents a Q-factor algorithm that optimizes scheduling decisions dynamically. The authors employ a variable neighbourhood search to explore the solution space and identify the most effective scheduling method. By using \ac{rl}, the authors determine the optimal parameters for rescheduling processes in response to changes in the environment. This approach is designed to address real-world challenges in manufacturing. The results demonstrate the significance of their method for a dynamic job shop environment, as the optimal strategies are also updated dynamically. 

By employing Google DeepMind's \ac{dqn} agent algorithm present a successful application of \ac{rl} to production scheduling. Their framework consists of multiple \ac{dqn} agents that cooperate and learn to achieve the defined objectives, resulting in self-organized, decentralized manufacturing systems that are capable of adapting to a dynamic manufacturing environment. After a short training phase the system presents scheduling solutions that are on par with solutions based on expert knowledge. Even though the approach cannot beat heuristics, this research represents a significant step towards the application of \ac{ai} to real-world industrial processes and intelligent production systems.

Further developing both Gabel and Riedmillers \cite{Gabel.2012} and Lillicrap et al.'s approach \cite{Lillicrap.2015}, in 2020 Liu et al. \cite{Liu.2020} utilized an actor-critic \ac{drl}-architecture to approach the \ac{jssp} as a sequential decision-making problem. The model consists of an actor network and a critic network, including convolution layers and fully connected layers. The actor network learns how to act in a dynamic environment, while the critic network evaluates these actions. This approach is effective in managing unexpected events such as machine breakdowns, additional orders or material shortages that may interrupt the production process. It offers more robust and efficient scheduling solutions while still producing comparative solutions for smaller benchmark problems, outperforming traditional dispatching rules and executing almost as fast as simple dispatching rules. With increasing sizes of the instances, the performance eventually declined.

Han and Yang \cite{Han.2020} deal with increased complexities and uncertainties of \ac{jssp}s by proposing a duelling double \ac{dqn} with priority replay. This \ac{drl}-framework combines the advantages of real-time response and flexibility of a deep convolutional neural network and \ac{rl} to dynamically respond to complex and changing production environments. Scheduling is seen as a sequential decision-making problem, where the scheduling states at each time step are expressed as multi-channel images. The action space is a combination of easy to execute heuristic rules. The duelling double \ac{dqn} is used to optimize learning through continuous interactions with the environment, resulting in a more accurate and stable learning performance in environments with high-dimensional action spaces and complex state evaluations. Experimental results show that this method achieves optimal solutions for small-scale problems and outperforms traditional heuristics for larger problems comparable to genetic algorithms but wasn't tested with a new dataset.

Zhang et al. \cite{Zhang.2020} developed a \ac{gnn} capable of solving problems regardless of their size. Like Han and Yang \cite{Han.2020}, they utilized a disjunctive graph to represent the state space of the \ac{jssp}. The \ac{gnn} interprets the disjunctive graph representation, allowing the system to generalize solutions across different scales of problems. The conducted experiments reveal that this approach enables the agent to learn high-quality \ac{pdr}s from basic features, outperforming existing \ac{pdr}s and performing well on much larger instances that were unseen in the training phase, although the generalized results fell short of being optimal.

Also, Park et al. \cite{Park.2021} use the combination of \ac{rl} and a \ac{gnn} to solve the \ac{jssp} by formulating the scheduling process as a sequential decision-making problem. They employ a \ac{gnn} for representation learning to encode the spatial structure of the \ac{jssp} into node features that are used to determine the optimal scheduling actions. The training of these components is conducted using \ac{ppo}. Experiments demonstrate that the \ac{gnn} approach can outperform traditional dispatching rules and other \ac{rl}-based approaches on various benchmark \ac{jssp}s. Furthermore, the framework can learn transferable scheduling policies that apply to new \ac{jssp} scenarios (in terms of size and parameters) without additional training, highlighting its adaptability and efficiency.

An overview of how to model the \ac{jssp} as a sequential decision process and how a \ac{drl} architecture can be applied to the \ac{jssp} is given by Wang et al. \cite{Wang.2021}. Just like Park et al. \cite{Park.2021} they employ a \ac{ppo} algorithm to reduce the complexity. The performance of the proposed model is compared with heuristic rules and meta-heuristic algorithms. It is shown that this approach not only produces comparative results but is also able to realize adaptive scheduling and shows a certain generalization capability when faced with unseen situations.

The approach of Oren et al. \cite{Oren.442021} doesn't specifically tackle the \ac{jssp} itself, but more generally NP-hard \ac{cop}s \cite{Garey.1976} that are found in job shop scheduling. The method is designed to work both online and offline, which is crucial for adapting to a dynamic production environment. A \ac{dqn} is used offline to learn optimal policies through a simulation environment. The graph representation of the states enables the system to handle varying problem sizes and configurations. These policies are applied online to optimize the decisions in real time. The combined approach utilizes available time for deliberations, effectively reducing the gap to dedicated \ac{cop} solvers.

Tassel et al. \cite{Tassel.482021} propose a new \ac{drl} algorithm that is specifically tailored for job shop scheduling tasks, using recent advances of \ac{drl} to handle the growing complexity of \ac{jssp}s. The authors developed a compact state space representation, along with a simple dense reward function that is closely related to the sparse make-span minimization objective of COP methods. Benchmark instances provided by Taillard \cite{Taillard.1994} show that their method finds solutions 11\% better make-span than the best dispatching rule on Taillard's instances, 10\% better than Han and Yang \cite{Han.2020} and around 18\% better than Zhang et al. \cite{Zhang.2020}.

Zhao and Zhang \cite{Zhao.2021} investigate the application of \ac{drl} in dynamic job-shop production control. A dynamic job shop is a type of job shop where the scheduling environment is not static but changes over time. This dynamic nature can stem from several factors, including the arrival of new jobs, machine breakdowns, variable processing times, or changes in job priorities. As intelligent manufacturing becomes more and more important in industrial production systems, they address the lack of an evaluation mechanism that can accurately measure the control efficiencies of different scheduling plans. The authors create a multi-objective optimization model for a production control system, and subsequently introduce \ac{drl} to it. This is followed by a proposal of a dynamic job shop production control method that is also based on \ac{drl} and the explanation of the collaboration strategy for multiple subsystems. Experiments proved that their approach is effective.

A comprehensive literature review on the applications of \ac{drl} in production systems is presented by Panzer and Bender \cite{Panzer.2022} in 2022. They discuss the challenges of modern production environments, such as the high level of complexity and the demand for high throughput, all while maintaining adaptability and robustness in case of variations in the process or unforeseen events. They highlight the increasing use of \ac{drl} to optimize production systems and the significant contributions and developments in the field that are improving the efficiency and flexibility of production processes. 89\% of the benchmarked implementations increase the scheduling performance, reached lower total tardiness, higher profits, or other problem-specific objectives. In the field of production scheduling, 67\% of the reviewed papers applied value-based algorithms.

With the growing interest in using \ac{rl} methods for production scheduling, it becomes increasingly challenging or even impossible to reproduce existing studies with the same degree of accuracy. To make the research more widely applicable and to exploit its strengths for industrial applications, Rinciog and Meyer \cite{Rinciog.2022} propose to standardize the approaches. They propose a framework for applying \ac{rl} in this context by modelling production scheduling as a \ac{mdp}. The standardization is done in three steps: The standardization of the description of production setups used in \ac{rl} studies is based on an established nomenclature. This is followed by the classification of \ac{rl} design decisions from existing publications. Finally, recommendations for a validation scheme that focuses on reproducibility and sufficient benchmarking are proposed.

A novel algorithm for improving the generalization capabilities and solution effectiveness of a \ac{drl} agent that solves \ac{jssp}s is proposed by Vivekanandan et al. \cite{Vivekanandan.2023}. The authors introduce a new method called Order Swapping Mechanism to achieve better generalized learning. By using a set of known benchmark instances \cite{Taillard.1994} they compare their results with the work of other groups \cite{Han.2020}\cite{Zhang.2020}\cite{Tassel.482021} that used the same benchmark instances and demonstrate that this approach outperforms previous methods. The results demonstrate that the agent does not outperform the approach of Tassel et al. \cite{Tassel.482021}, yet it does provide a size-dependent generalization. It outperforms the \ac{pdr} based \ac{drl} approach of Zhang et al. \cite{Zhang.2020} and performs similarly to other state-of-the-art \ac{drl} algorithms.

Serrano-Ruiz et al. \cite{SerranoRuiz.2024} present a method for scheduling in a quasi-realistic job shop environment. They create a digital twin of the job shop model as a \ac{mdp} and use \ac{drl} for optimization. Their approach uses a deterministic framework for formulation and implementation and is validated by comparison with known heuristic priority rules. Experiments show that the model not only captures the benefits of heuristic rules but also leads to a more balanced performance across various indicators, outperforming traditional heuristic methods.

The analyzed papers demonstrate that the \ac{jssp} can be effectively solved by using various \ac{drl} approaches. However, most of the papers are based on simplified models with little relevance to the reality of production. In this reality, a large number of factors play a role, the quantification of which is sometimes difficult and can have a significant influence on the effectiveness or success of the modelling. The decisive factors and their influence on production planning are described in more detail in Section \ref{cha:Method}.

\section{Methodology}\label{cha:Method}
\subsection{Necessity for an Extended Approach}\label{cha:Necessity}
Despite the existence of various \ac{rl} approaches proposed by different researchers with the intention of solving generic \ac{jssp}s in industry-relevant problem sizes, it remains a challenge to translate the complexity of a real-world production environment into generic \ac{jssp}s. The following complexities are identified throughout this work:

\begin{enumerate}
    \item Generic \ac{jssp}s typically involve machines that process various jobs without any specific machine setup. It is therefore necessary to reduce the time required for machine setup to a minimum, as this can otherwise result in a significant loss of production time.
    \item Jobs are typically defined by a machine sequence and a fixed processing time. In the case of varying batch sizes, the processing times may be approximately linearly dependent on the batch size. Previous approaches did not consider varying batch sizes or processing times. With an enhanced generalization capability, contemporary \ac{drl} agents should be capable of accommodating varying processing times.
    \item The field of intralogistics is not included in the scope of a generic \ac{jssp}. The transportation times between different machines or production facilities can be considerable, often taking several minutes, and therefore play an important role in the scheduling process.
    \item In a real-world production environment, a variety of storage spaces can be found to buffer inconsistencies in production processes or to increase production flexibility. The dimensions of these buffering zones were not considered in previous approaches. An overfilled buffer zone may impede the entire production process and thus necessitates consideration in the scheduling process.
    \item Deadlines are frequently absent from models, even though they are an essential component of ensuring the timely delivery of products.
\end{enumerate}

These simplifications make the \ac{jssp} easier to model mathematically, but real-world manufacturing systems often involve complex scheduling challenges that are not captured by these simplifications. In order to create a training environment that closely resembles the complex processes of a real production environment, it is necessary to develop an extended approach with a higher level of information detail.

\subsection{Research Hypothesis and Objective}
Based on a comprehensive analysis of an existing furniture factory, a concept for the implementation of \ac{rl} in production planning is proposed. In order to formulate a general concept that is valid for a broad industry, the following constraints are given:

\begin{itemize}
    \item The production is set up as a Job Shop
    \item The machines produce products in batches
    \item The batches are manufactured on a recurring basis, but the production is frequently switched to enable the production of a wider range of products
    \item The overall range of products doesn't change drastically after training, as this would require a re-training of the agent. Minor changes like "color changes" would not disrupt the production process.
\end{itemize}

\subsection{The Model}
Generic \ac{jssp}-models are typically described by $n$ jobs $J = \{J_1, J_2, ..., J_n$\}, where each job has $m$ operations $O$ ${(J_i = \{O_{i1}, O_{i2}, ..., O_{im}\})}$ to be processed on $m$ machines $M$ = \{$M_1$, $M_2$, $...$, $M_m$\}. Each operation has its designated processing machine and time $d_{ij}$. This model is extended by the introduction of the following elements:

\paragraph{Job Volumes}
    The volumes of the jobs before and after each operation are quantified and mapped. With knowledge of the volumes of each job at any given point in the production process, the required storage spaces can be estimated with greater precision. This information is used to forecast and analyze the utilization of the buffers.
    
\paragraph{Buffers}
    In a production system, storage areas are distributed across the shop floor and are used to hold materials, unfinished, or finished goods between different stages of production or between production and shipping. The primary purpose of these areas is to absorb variability in the production process, including fluctuations in demand, supply disruptions, machine breakdowns, or other unforeseen circumstances that may impact the production schedule. These storage areas are defined as buffers and are used to store the jobs before being processed at a machine. Each buffer is characterised by its capacity, which may be expressed in various units, including storage volume, pallet storage capacity, or any other meaningful unit. Buffers serve to absorb variability in the production process, enabling a more flexible production, preventing bottlenecks and thereby smoothing out the production flow.
 
\paragraph{Quantity Factor $\delta$}
    The processing time of each operation is subject to significant variation due to fluctuations in batch sizes, which are quantified using the quantity factor $\delta$. The quantity factor is employed to determine the duration for which a machine is occupied in processing a component, particularly when batch sizes exhibit considerable variability. The total processing time of an operation is calculated as follows:
    
    \begin{align}
    \text{total processing time of an operation}&= \delta \cdot d_{ij}&& \text{with $d_{ij}$ as the processing time of one single element of an operation}
    \end{align}\label{tot.proc.time}
    
    This methodology enables the calculation of the processing time of a job on a machine based on the actual batch size of a job.

\paragraph{Transportation Times $t$}
    In order to describe the intralogistic processes, transportation times are introduced. These figures represent the time required to transport a job from one machine to the next. In the majority of cases, the transportation times between two points are symmetrical, as the required transportation time is independent of the direction of travel. Table \ref{tab:transportation} illustrates an example of transportation times.
    
\begin{table}[ht]
	\caption{Example for transportation times (mock-up data)}
	\centering
	\begin{tabular}{ccccc}
		\toprule
		Transportation Time&to Machine 1& to Machine 2&to Machine 3\\
		\midrule
		from Machine 1  &  0   &   10  &  15\\
		from Machine 2  &  10  &   0   &  15\\
		from Machine 3  &  15  &   15  &   0\\
		\bottomrule
	\end{tabular}
	\label{tab:transportation}
\end{table}

\paragraph{Machine Setup Times $s$}
    A machine that is involved in the production of various jobs may require different setups in order to perform the specific operations. These setup times are considered, when the production requires a switch from one setup to another. Table \ref{tab:symm} illustrates an example of symmetric setup times, where the individual setup times are mostly independent of the previous setup. The neutral setup represents a neutral state of the machine, for example a saw without an installed saw blade.
    
\begin{table}[ht]
	\caption{Example for symmetric machine setup times (mock-up data)}
	\centering
	\begin{tabular}{ccccc}
		\toprule
		Machine 1     &   to neutral Setup    &   to Setup 1  &   to Setup 2  &   to Setup 3\\
		\midrule
		from neutral Setup  &  0   &   4   &   4   &   4   \\
		from Setup 1        &  4   &   0   &   8   &   8   \\
		from Setup 2        &  4   &   8   &   0   &   8   \\
            from Setup 3        &  4   &   8   &   8   &   0   \\
		\bottomrule
	\end{tabular}
	\label{tab:symm}
\end{table}

    In certain instances, the process of setting up a machine may take longer than the process of dismantling it. This phenomenon is indicated in Table \ref{tab:asymm}, which represents an example of asymmetric setup times. The individual setup times are significantly influenced by the previous setup. The dismantling of a machine to a neutral setup is shorter than setting up any other setup. However, switching directly between setups reveals synergies that shorten the overall setup time.
    
\begin{table}[ht]
	\caption{Example for asymmetric machine setup times (mock-up data)}
	\centering
	\begin{tabular}{ccccc}
		\toprule
		Machine 2     &   to neutral Setup    &   to Setup 1  &   to Setup 2  &   to Setup 3\\
		\midrule
		from neutral Setup  &  0   &   9   &   7   &   8   \\
		from Setup 1        &  5   &   0   &   10  &   13  \\
		from Setup 2        &  5   &   8   &   0   &   6   \\
            from Setup 3        &  5   &   8   &   7   &   0   \\
		\bottomrule
	\end{tabular}
	\label{tab:asymm}
\end{table}

    In automated CNC machines, different jobs may have different machine setups, which can be achieved without the need for a physical setup change. Instead, a virtual setup change can be implemented with a simple mouse click. Consequently, the actual setup times do not need to be calculated. Table \ref{tab:asymmWO} illustrates an example of this phenomenon, demonstrating the transition between Setup 1 and Setup 2.

\begin{table}[ht]
	\caption{Example for asymmetric machine setup times, partially without physical setup change (mock-up data)}
	\centering
	\begin{tabular}{ccccc}
		\toprule
		Machine 3     &   to neutral Setup    &   to Setup 1  &   to Setup 2  &   to Setup 3\\
		\midrule
		from neutral Setup  &  0   &   4   &   7   &   5   \\
		from Setup 1        &  2   &   0   &   0   &   7   \\
		from Setup 2        &  2   &   0   &   0   &   6   \\
            from Setup 3        &  2   &   6   &   8   &   0   \\
		\bottomrule
	\end{tabular}
	\label{tab:asymmWO}
\end{table}

\paragraph{Overall Processing Time $T_{total}$}
    With the introduction of the quantity factor $\delta$, the transportation time $t$ and the setup time $s$, the overall processing time of an operation can be calculated as follows:
    
    \begin{equation}\label{eq:totproctime}
        T_{total} = \delta \cdot d + t + s
    \end{equation}

\paragraph{Deadlines}
    The deadline is the latest point in time at which the production of a job must be completed. The deadline can refer to a shipping date or an upcoming maintenance schedule for a machine or a complete production line. Failure to meet the deadline can result in several consequences, making it an important factor to consider in production planning.

This extended approach allows for the acquisition of more detailed information. An example of the resulting operations is illustrated in Table \ref{tab:Operations}.
Taking these factors into account, a training environment can be created that bridges the gap between reality and model.

\begin{table}[ht]
	\caption{List of operations of the exemplary \ac{jssp} (mock-up data)}
	\centering
	\begin{tabular}{ccccccc}
		\toprule
		Operation name     &   Machine    &   Machine Setup  &   Machining time  &   Quantity & Volume & Deadline\\
		\midrule
		$O_{11}$ &  M1  &   $s_1$   &   0,04  & 400 & 30 & - \\
		$O_{12}$ &  M2  &   $s_1$   &   0,08  & 400 & 20 & - \\
		$O_{13}$ &  M3  &   $s_1$   &   0,06  & 400 & 15 & 120\\
            $O_{21}$ &  M1  &   $s_2$   &   0,12  & 100 & 10 & - \\
            $O_{22}$ &  M3  &   $s_2$   &   0,14  & 100 &  8 & - \\
            $O_{23}$ &  M2  &   $s_2$   &   0,13  & 100 &  5 & 110\\
            $O_{31}$ &  M1  &   $s_3$   &   0,06  & 400 & 20 & - \\
            $O_{32}$ &  M2  &   $s_3$   &   0,04  & 400 & 15 & - \\
            $O_{33}$ &  M3  &   $s_3$   &   0,06  & 400 & 10 & 100\\
		\bottomrule
	\end{tabular}
	\label{tab:Operations}
\end{table}

\subsubsection{Environment Outline}\label{example}
The job shop environment can be implemented using toolkits such as OpenAI Gym and libraries such as StableBaselines3, which are specifically designed for developing and comparing reinforcement learning algorithms. In the training environment, an agent learns to solve the problem and optimize its policy parameters by interacting with the environment through actions.

The following example \ac{jssp} is considered:

Three jobs $J$ = \{$J_1, J_2, J_3$\} are to be produced on three machines $M$ = \{$M_1, M_2, M_3$\} with a machine sequence of
\begin{align*}
    J_1 = \{M1, M2, M3\}\\
    J_2 = \{M1, M3, M2\}\\
    J_3 = \{M1, M2, M3\}
\end{align*}

The batch sizes of the jobs $J$ are 
\begin{align*}
    J_1 = 400\,\mathrm{pcs}\\
    J_2 = 100\,\mathrm{pcs}\\
    J_3 = 400\,\mathrm{pcs}
\end{align*}

The required machine setups for the operations are
\begin{align*}
    O_{11} = s_1 \quad O_{12} = s_1 \quad O_{13} = s_1\\
    O_{21} = s_2 \quad O_{22} = s_2 \quad O_{23} = s_2\\
    O_{31} = s_3 \quad O_{32} = s_3 \quad O_{33} = s_3
\end{align*}

and the machining times for a single one element of each job are
\begin{align*}
    d_{11} = 0.04\,\mathrm{min} \quad d_{12} = 0.08\,\mathrm{min} \quad d_{13} = 0.0625\,\mathrm{min}\\
    d_{21} = 0.12\,\mathrm{min}\, \quad d_{22} = 0.14\,\mathrm{min} \quad d_{23} =\, 0.13\,\,\mathrm{min}\,\,\,\\
    d_{31} = 0.0625\,\mathrm{min} \quad d_{32
} = 0.04\,\mathrm{min} \quad d_{33} = 0.0625\,\mathrm{min}
\end{align*}

With this information and the assumption that Job 3 is already located at Machine 1, the overall processing time of operation $O_{31}$ can be calculated as follows:
    \begin{align}\label{eq:totproctimeO31}
    \begin{split}
        T_{total_{O_{31}}} &= \delta \cdot d + t + s\\
                           &= 400    \cdot 0.0625\,\mathrm{min} + 0\,\mathrm{min} + 4\,\mathrm{min}\\
                           &= 29\,\mathrm{min}
    \end{split}
    \end{align}

The volumes of the jobs change with every operation
\begin{align*}
    V_{11} = 30\,\mathrm{m^3} \quad V_{12} = 20\,\mathrm{m^3}       \quad V_{13} = 15\,\mathrm{m^3}\\
    V_{21} = 10\,\mathrm{m^3} \quad V_{22} =  \,\,\,8\,\mathrm{m^3} \quad V_{23} = \,\,\,5\,\mathrm{m^3}\\
    V_{31} = 20\,\mathrm{m^3} \quad V_{32} = 15\,\mathrm{m^3}       \quad V_{33} = 10\,\mathrm{m^3}
\end{align*}

Table \ref{tab:Operations} shows a corresponding list of operations.

A buffer is located in front of every machine. In order to be processed on a machine, a job needs to be stored in the respective buffer first. The buffer capacities are
\begin{align*}
    \mathrm{B}1 = 60\,\mathrm{m^3}\\
    \mathrm{B}2 = 43\,\mathrm{m^3}\\
    \mathrm{B}3 = 30\,\mathrm{m^3}
\end{align*}

\begin{figure}[ht]
    \centering
    \includegraphics[width=\textwidth]{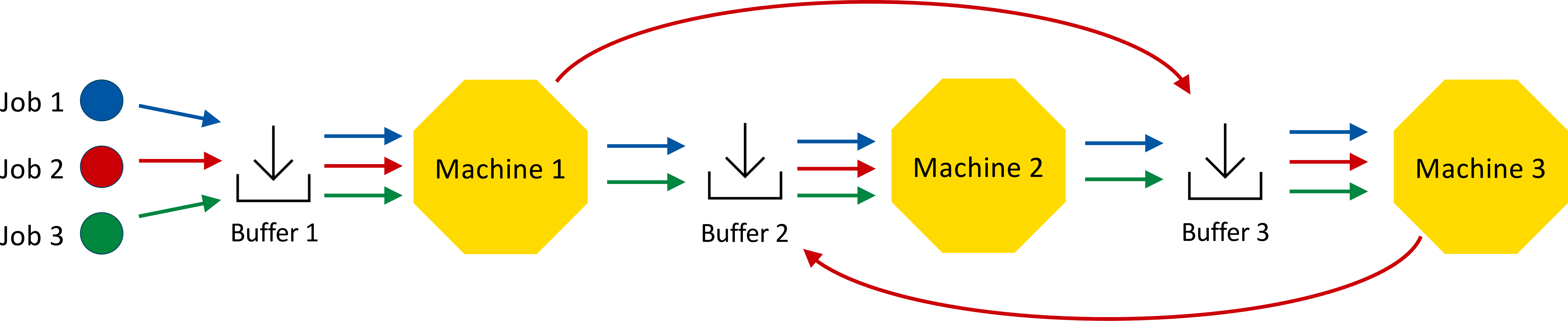}
    \caption{Production mapping of exemplary JSSP}
    \label{fig:ProdMap}
\end{figure}

With this information, the production flow can be mapped as shown in Figure \ref{fig:ProdMap}.

To ensure the agent's training closely resembles reality, the training environment is designed with specific constraints:\cite{Pinedo.2012}\cite{Bazewicz.2001}\cite{Vivekanandan.2023}

\begin{enumerate}
    \item Machines are limited to processing one job at a time
    \item The processing of an operation cannot be interrupted
    \item There are no precedence constraints between operations of different jobs
    \item Each job has a fixed machine sequence
    \item Each operation requires a specific machine setup
    \item Machines need to be set up accordingly before processing a job.
\end{enumerate}

\paragraph{Time Step Transition}\cite{Vivekanandan.2023}
In accordance with these constraints, the environment treats operations as atomic, implying that they cannot be interrupted, even if they consist of multiple individual elements (based on their batch size). Once a machine has been assigned to a job, it becomes occupied and is no longer available for assignment to other jobs. In the provided example, jobs $J_1$, $J_2$, and $J_3$ all have Machine $1$ as their first machine in their machine sequence. Once the agent has selected Machine $1$ for one of the assigned jobs, no further actions are necessary, and the agent can proceed to the next time step. In order to ensure optimal agent training, time steps and machine assignments are determined based on the eligibility of the operations. At each time step, the agent identifies all eligible operations $O_{ij}$, based on the predefined job order of the problem scenario. The eligibility criteria for an operation depend on the job order and the current status of the machines. This approach facilitates targeted learning by presenting the agent with relevant and actionable decision points, thereby simplifying the problem-solving process within the computational framework.

Due to its sequential nature, the process of assigning a machine to a job can be modelled as a \ac{mdp}.

\subsubsection{Action Space}
The environment is controlled by a single discrete action space, through which the agent selects the appropriate jobs for processing on a given machine at each time step. The agent's choices are limited to the subset of available jobs and machines, ensuring that the agent's decisions are both relevant and feasible in the context of the current operational parameters and machine availability. The actions taken by the agent modify the environment and change its state.

\subsubsection{Observations}\label{observations}
The decision to assign a machine to a job is based on a set of observations that is provided to the agent. Each observation represents the current state of the environment. It is updated at each time step and holds the following information:

\textbf{(1) machine info}\\
    A matrix of dimensions $3 \times m$, which contains information about the currently processed jobs, the progress of the operations, and the current machine setup. Given that $m$ represents the number of machines in the environment, each column of the matrix represents a machine. The first line contains information about the currently processed jobs, the second line contains information about the remaining operation time and the third line contains information about the current machine setup.

\begin{equation*}
(1) t_0=\begin{bNiceMatrix}[last-row]
0&0&0\\
0&0&0\\
\aoverbrace[0lDr0]{\text{\small{Machine 1}}} & \aoverbrace[0lDr0]{\text{\small{Machine 2}}} & \aoverbrace[0lDr0]{\text{\small{Machine 3}}} \\
\end{bNiceMatrix}
\end{equation*}

\textbf{(2) job info}\\
    A matrix of dimensions $2 \times n$, containing information about the current job volumes and the remaining time until the deadline is reached. With $n$ representing the number of jobs in the environment, each column represents a job. The first line contains information about the current job volume, while the second line contains information about the remaining time until the deadline is reached.

\begin{equation*}
(2) t_0=\begin{bNiceMatrix}[last-row]
30&10&20\\
120&110&100\\
\aoverbrace[0lDr0]{\text{\small{Job 1}}} & \aoverbrace[0lDr0]{\text{\small{Job2}}} & \aoverbrace[0lDr0]{\text{\small{Job 3}}} \\
\end{bNiceMatrix}
\end{equation*}

\textbf{(3) buffer info}\\
    A $b$-dimensional vector represents the capacity status of the buffers in the environment with $b$ representing the number of buffers in the environment. This vector can reflect the utilization of the buffers in units or in percent, depending on what is most appropriate for the model.

\begin{equation*}
(3) t_0=\begin{bNiceMatrix}[last-row]
60&0&0\\
\aoverbrace[0lDr0]{\text{\small{Buffer 1}}} & \aoverbrace[0lDr0]{\text{\small{Buffer 2}}} & \aoverbrace[0lDr0]{\text{\small{Buffer 3}}} \\
\end{bNiceMatrix}
\end{equation*}

These three observations provide an overview of the status of the entire environment. Additional information, such as machine availability, can be extracted from these observations. If a machine has no job assigned to it, it is considered to be idling and available.

\subsubsection{Transition Probability Function}
Transition probabilities represent the likelihood of transitioning from one state to another after taking a given action. The transition probabilities are estimated based on the experiences of past actions and resulting observations in the training environment. They are then continuously adapted and honed in the training phase. This function is capable of capturing the dynamics of the system under different observations.

\subsubsection{Reward}
The reward function is employed to quantify the system performance and guide the decision-making process. It must be tailored closely to the goals of the specific industrial application, as it is highly sensitive. The reward corresponds to the scheduling goal and defines the specialization of the agent. In order to maintain a high level of learning efficiency, it is necessary for the agent to be rewarded in a densely manner. This implies that the agent is rewarded for actions that are specific to the JSSP, such as achieving a shorter overall processing time. Additionally, the agent may be rewarded for actions that are specific to its industrial application. One method of enhancing the learning process is to provide the agent with a negative reward for unwanted actions, such as overfilling a buffer or failing to meet a job's deadline. Identifying the optimal reward function is an iterative process that must be conducted on an individual basis.

\paragraph{Discount Factor $\gamma$}
As with the reward function, the discount factor must be set according to the specific application in question. The discount factor between 0 and 1 determines the relative importance of future rewards in the decision-making process. A discount factor of 0 indicates that the agent is short-sighted and only considers immediate rewards, while a discount factor close to 1 implies that the agent values future rewards to a similar extent as immediate rewards.

In order to identify the optimal policy, the application of \ac{ppo} balances the trade-off between policy improvement and stability.

\section{Integration in the Furniture Industry}\label{cha:Integration}
In a general JSSP, each job is associated with a unique set of tasks that must be completed in a specific order. This concept can also be applied to the production of furniture, where articles are composed of several components that must be processed in a specific sequence on various machines. Consequently, each article corresponds to a group of jobs, with each component considered being a separate job. In order to determine this set of jobs, it is necessary to break down each article into its various components and determine their respective machining sequences. The identification of production information, such as the individual machining sequence, volume alterations, and processing duration for each component at every production step, is typically obtained from the so-called \ac{bom}, which is a commonly utilized document in the industry. The level of detail provided by the \ac{bom} determines the comprehensiveness of the production information. Obtaining such information is indispensable for the development of optimized production planning. Given the intricate nature of furniture production systems, the implementation of \ac{drl}-supported production planning necessitates a comprehensive analysis of the factory in question. Consequently, the presented solutions thereby may not be readily transferable to all furniture production facilities, as each facility will require a solution that is precisely tailored to the factory and its individual scheduling goals. This chapter presents an example of how a \ac{rl}-agent can be employed for production scheduling in the furniture industry, where components are manufactured in batches in a job shop environment. For that purpose, the elements of the extended model that is presented in Chapter \ref{cha:Method} may be represented as follows:

\paragraph{Jobs}
    A piece of furniture is typically constructed from multiple components, which are then assembled or packaged at a designated station in order to create the final product. In the context of the job shop analogy, each component represents a distinct job with its own machine sequence and processing time for each operation. Therefore, the furniture articles are broken down into individual components. For each component, the processing sequence on the respective machines, the processing time, and the machine setups are mapped. If no further processing occurs within the factory, purchased parts may be excluded from this analysis. Each job is then defined by its machine sequence, machine setup, total processing time (cf. \ref{tot.proc.time}), deadline, and its volume, which undergoes change throughout the production process.

\paragraph{Job volumes}
    It is common in furniture production, particularly for solid wood furniture, to undergo significant volume changes throughout the production process. The volume of the raw material at the beginning of the process may be up to five times greater than that of the end product. In order to calculate the space requirements of each job at any point during production, it is necessary to record the volume change in each process step.
    
\paragraph{Buffers}
    Each buffer on the shop floor is assigned to a machine and located in front of it in the process flow. Buffers are treated similarly to machines, but they can store several jobs simultaneously and have no processing time. The capacity of these buffers is specified in terms of the number of storage spaces for pallets or by volume.
    
\paragraph{Machines}
    Another aspect of the environment is the machinery used for production. Each machine is mapped in detail, including its features, processing times, possible setups, and setup times required for each operation.

\paragraph{Quantity Factor $\delta$}
    The quantity factor $\delta$ is employed to modify batch sizes, which have a significant impact on processing time and, consequently, the make span of a job. Training the agent with varying batch sizes enhances its generalization capability, enabling it to handle varying job sizes. It has been observed that the production of furniture does not always occur in consistent batch sizes. The quantity factor helps determining the duration of a machine's utilization in the processing of a component.

\paragraph{Transportation Times}
    The transportation times between machines and buffers in the production are mapped. This includes a full analysis and documentation of the transportation times between the machines on the shop floor. With this information, the overall completion time of an order can be determined.

\paragraph{Deadlines}
    Each job is described with a deadline that specifies the latest point in time by which its processing must be completed.
    
\paragraph{Environment}
    The production environment is mirrored as closely as possible by the training environment, which comprises jobs, machines, and buffers. The agent is trained to set up machines, assign jobs to machines, and prevent buffers from overflowing, all while maintaining the correct machine order for each job and meeting the production deadlines.
    
\paragraph{Action Space}
    The agent is controlling the environment in a single discrete action space. At each time step suitable machines and buffers are determined for the available jobs.
    
\paragraph{Observations}
    The observations provided to the agent are (1) machine info, (2) job info, and (3) buffer info (cf. \ref{observations}).
    
\paragraph{Reward}
    The reward function guides the decision-making process and thus must be tailored to the specific optimization goal. It needs to include rewards for correctly setting up a machine for an according job, assigning machines to jobs in the correct machine sequence of a job and storing jobs in buffers before assigning them to a machine. Additionally, the reward function contains rewards for the specific optimization goals, such as low buffer levels, reducing make span times, or achieving other, industry-specific objectives.

\subsection{Integration Strategy}
The integration of an \ac{rl} agent must be compatible with existing production and planning systems and be able to be seamlessly integrated without disrupting ongoing production. Once the training environment has been established and the agent has been trained in accordance with the desired optimization goals, it can then be utilized to support production planning.
Two principal concepts may be distinguished with regard to the integration: episodic and continuous production planning. The decision to utilize either episodic or continuous planning depends on various factors, including the nature of the production process, the degree of variability in orders, and the scheduling system's required flexibility.

\subsubsection{Episodic Planning}
\begin{figure}[ht]
    \centering
    \includegraphics[width=15cm]{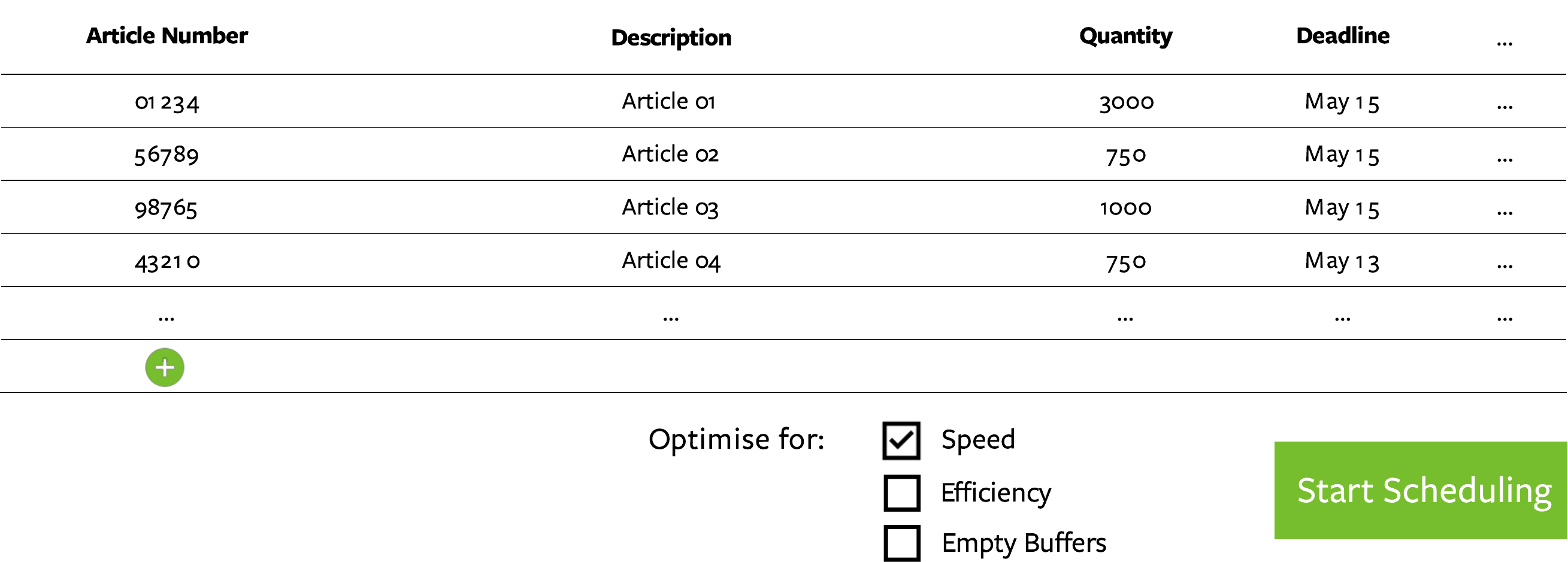}
    \caption{Exemplary layout of a production planning dashboard}
    \label{fig:Dashboard}
\end{figure}

In companies with low levels of automation and networking, episodic planning represents an appropriate approach. It may be employed when orders are processed in batches or when the production line is configured to manufacture specific types of furniture for a defined period before transitioning to a different type. The integration of an episodic planning system is less complex than in a continuous planning system (cf. \ref{cha.continuous planning}), as it does not require interfaces with existing production systems. The planning process is divided into distinct episodes, each with a clear beginning and end. These episodes may be defined as the production of a week or the production of an individual order from a specific client. Each episode consists of a limited number of steps and actions, making it easier to predict and evaluate outcomes. The orders are entered into a dashboard, which includes information on quantities and deadlines. Alternatively, a specific time period may be selected for scheduling. Figure \ref{fig:Dashboard} illustrates an exemplary dashboard, which displays orders, quantities, and deadlines. The system may offer the users a range of optimization goals to choose from, or alternatively, suggests different schedules based on various models. This enables production planners to respond effectively to the dynamic changes of the production environment.

Once the orders have been entered into the dashboard, the agent starts with the scheduling process. With the \ac{jssp} example provided in chapter \ref{example}, the initial observations at $t_0$ are as follows:

\begin{equation*}
(1){t_0}= \begin{bmatrix}
0 & 0 & 0\\
0 & 0 & 0\\
0 & 0 & 0
\end{bmatrix}
\end{equation*}    

\begin{equation*}
(2){t_0}= \begin{bmatrix}
30  & 10  & 20\\
120 & 110 & 100
\end{bmatrix}
\end{equation*}  

\begin{equation*}
(3){t_0} = \begin{bmatrix}
60 & 0 & 0
\end{bmatrix}
\end{equation*}

\begin{figure}[ht]
    \centering
    \includegraphics[width=15cm]{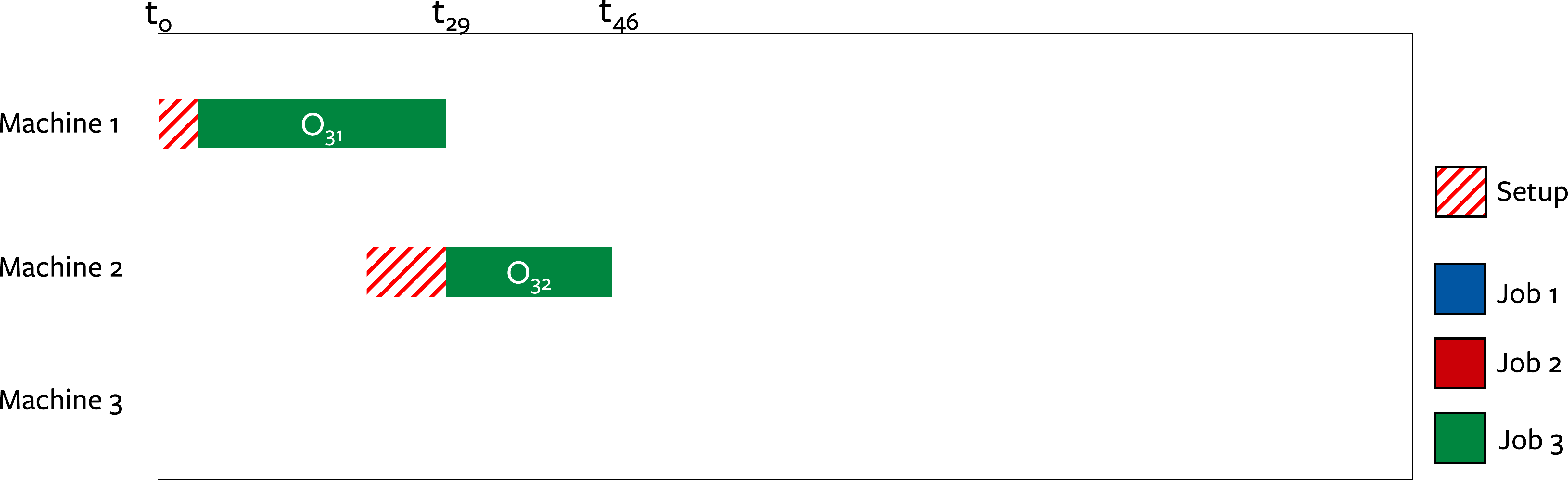}
    \caption{Scheduling with a trained agent at $t_0$}
    \label{fig:Schedule_t_0}
\end{figure}

At the initial time step $t_0$, all machines are in neutral setups, idling, and thus available for use. All three jobs require Machine 1 as their first machine in the sequence. The agent assigns Job 3 to Machine 1, sets up the machine for the job, and creates a new time step $t_{29}$, for when the processing of Job 3 will be finished (cf. Formula (\ref{eq:totproctimeO31})). Machine 2 is set up to take over Job 3 at the new time step $t_{29}$, that will eventually be finished at time step $t_{46}$, as illustrated in Figure \ref{fig:Schedule_t_0}.

Since there are no other viable operation at $t_0$, the agent skips to the next time step $t_{29}$, where it is provided with a new, updated set of observations:

\begin{equation*}
    (1){t_{29}}= \begin{bmatrix}
 0 & 0 & 0 \\
 0 & 0 & 0 \\
 s_3 & s_3 & 0 
\end{bmatrix}
\end{equation*}    

\begin{equation*}
    (2){t_{29}}= \begin{bmatrix}
30  & 10  & 15\\
91  & 81  & 71
\end{bmatrix}
\end{equation*}  

\begin{equation*}
    (3){t_{29}} = \begin{bmatrix}
40 & 15 & 0
\end{bmatrix}
\end{equation*}

\begin{figure}[ht]
    \centering
    \includegraphics[width=15cm]{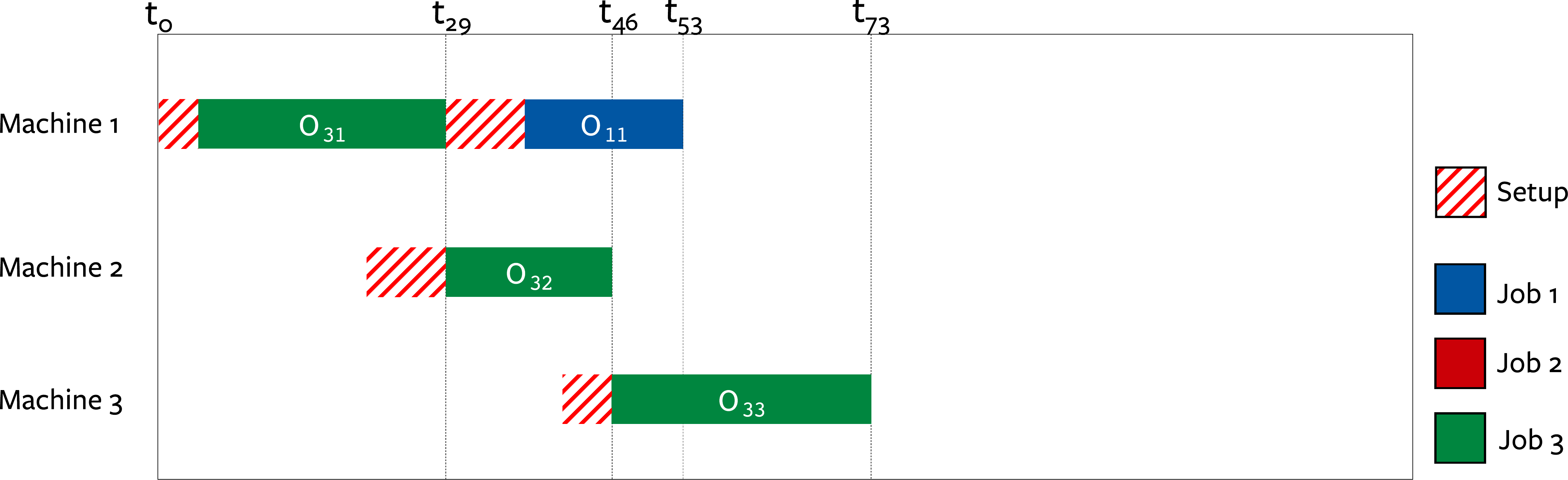}
    \caption{Scheduling with a trained agent at $t_{29}$}
    \label{fig:Schedule_t_29}
\end{figure}

Based on this new set of observations, the agent decides at $t_{29}$ to assign Job 1 to Machine 1, set it up accordingly, and create a new time step $t_{53}$ for the end of processing Job 1 on Machine 1. Furthermore, Machine 2 is assigned with Job 3, while Machine 3 is set up to process Job 3 after it is done on Machine 2 at $t_{46}$. This is shown in Figure \ref{fig:Schedule_t_29}.

\begin{figure}[ht]
    \centering
    \includegraphics[width=15cm]{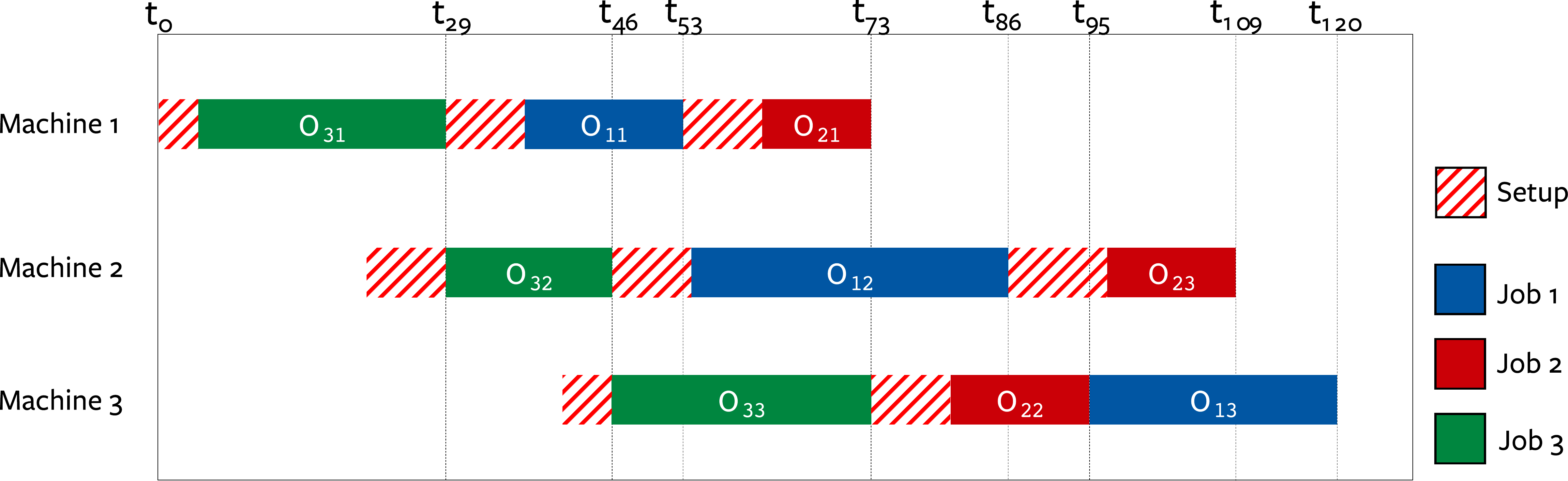}
    \caption{Completed scheduling}
    \label{fig:Schedule_full}
\end{figure}

These time step transitions are repeated until no operations are left for processing and the production is completed. The finished schedule is shown in Figure \ref{fig:Schedule_full}. After completion, the system resets, allowing for evaluation and optimization of each episode independently.

\paragraph{Limitations of Episodic Planning}
In a real-world production environment, a number of factors may influence the production planning process, including unforeseen events such as machine breakdowns, delivery delays, or changes in customer requirements. As the episodic planning process is conducted in advance, it is not possible to incorporate unforeseen events into the scheduling process.
The introduction of a new product to the portfolio necessitates the re-training of the agent. Even an agent with high generalization capabilities is unable to predict the correct order of operations for the components of a new furniture article. A trained agent is constrained to the specific set of jobs for which it was trained.

\subsubsection{Continuous Planning}\label{cha.continuous planning}
\begin{figure}[ht]
    \centering
    \includegraphics[width=6cm]{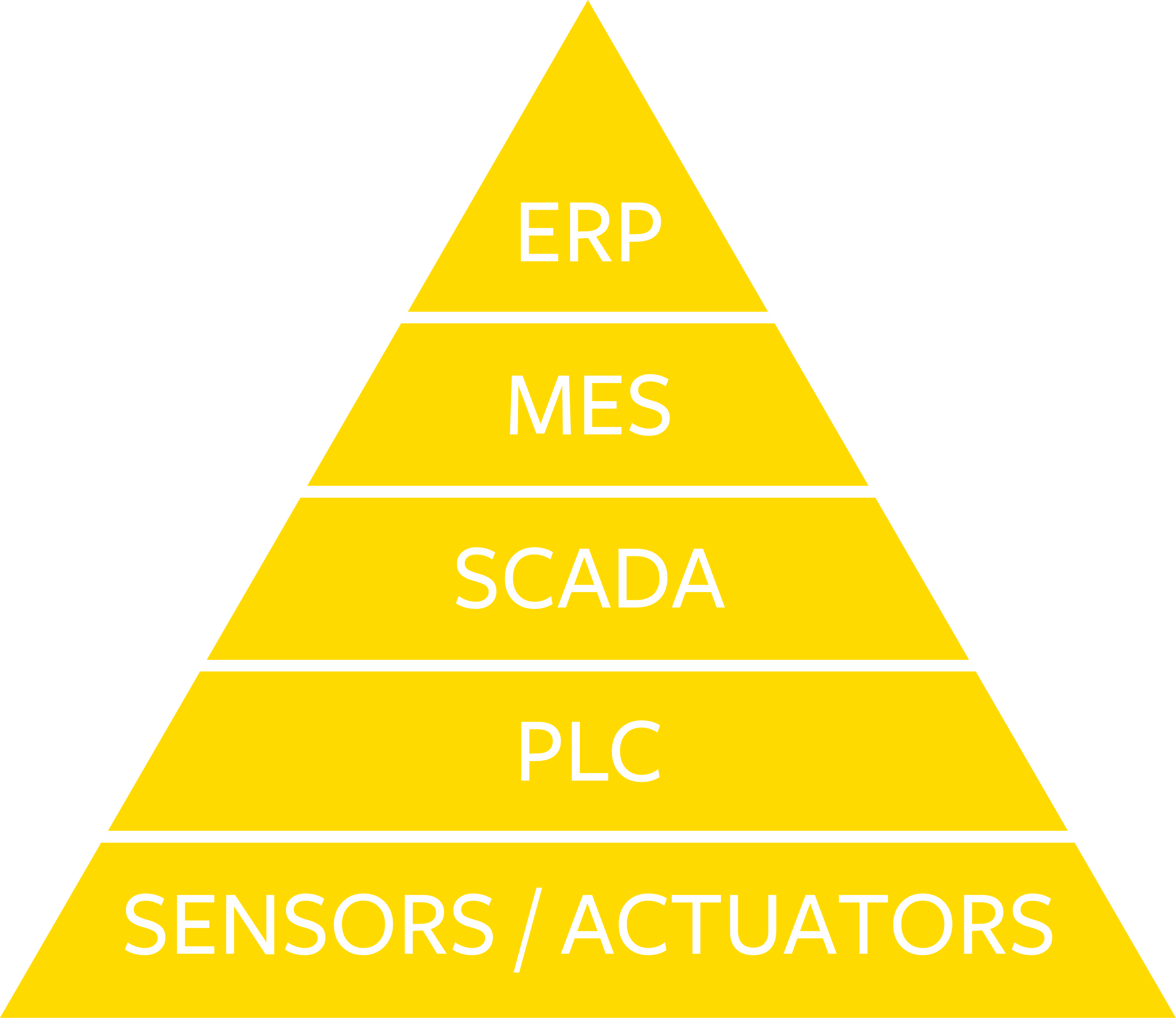}
    \caption{Automation Pyramid}
    \label{fig:auto_pyra}
\end{figure}

In a continuous planning approach, the agent is fully integrated into the production system and plays an active role in the scheduling process. This approach is suitable for highly networked and automated plants where production is subject to significant variations in orders, requiring production schedules to be adjusted on the fly to accommodate new orders, changes in design, or material availability. This integration comes at the cost of a significantly higher level of complexity: The agent is situated between the \ac{erp}-system and the \ac{mes} (see Figure \ref{fig:auto_pyra}).
It processes real-time production data from both systems and adjusts the production accordingly. The \ac{erp} system provides information regarding orders, material stocks, and delivery changes, while \ac{mes} provides machine data such as processing times, operation progression, machine breakdowns, or transport times of transport systems. Significant alterations to this data will trigger an event, for which the agent will recalculate the scheduling, analogous to a time step in episodic planning. The scheduling results are written back into the systems, where production planners can supervise and adjust the proposed scheduling, if necessary. Unlike the episodic approach, which only provides various scheduling suggestions for the planned episodes, the production planning is actively influenced by the agent. 

\paragraph{Limitations of Continuous Planning}
optimized production planning frequently requires the simultaneous consideration of multiple objectives, including a minimized lead time, maximized machine utilization and minimized inventory costs. These objectives may, however, be in conflict with one another. Despite the agent's full integration into the production systems, the choice of the optimization goal remains the prerogative of an experienced production planner, who sets and adjusts the goals according to current needs. The interface communication between the agent, the \ac{erp} system and the \ac{mes} requires a highly detailed customization, tailored to the specific needs and local conditions on site. As with the episodic approach, the continuous approach requires a retraining of the agent, when a new furniture article is entered into the system.

\subsection{Implementation Strategy}
The following procedure is proposed for the successful implementation of a \ac{drl}-supported production planning system:
\begin{enumerate}
    \item \textbf{preparation and planning:}
    A detailed examination of the interfaces between the systems points out if specific adapters need to be developed. Furthermore, the data flow required to exchange relevant information between the agent and the systems must be defined. This may include production schedules, machine statuses, order details, and other system-specific information.
    \item \textbf{goal definition:}
    Measurable results are achieved by the precise definition of the agent's goals and objectives. These goals may be the optimization of the throughput, the reduction of production costs or other specific goals.
    \item \textbf{definition of the state space:}
    This includes defining the relevant variables and factors that influence the state of the production system as well as defining the data flow, i.e. how the required information is received, transmitted, and processed by the agent. It is essential that the observations include all information relevant for the agent's decision-making process.
    \item \textbf{definition of the action space:}
    A precise definition of the actions that the agent can perform to influence the system's state. Examples of actions include the adjustment of machine setups, the assignment of machines and jobs or a simple no-op (no operation).
    \item \textbf{definition of the rewards and penalties:}
    In \ac{rl}, the agent is trained through the provision of feedback in the form of rewards and penalties. By clearly defining the reward function, the specialization of the agent can be specified. The reward definition should represent the goals defined at the beginning of this procedure. This may be an iterative process in the training phase. An example could be a reward for an increased throughput or a penalty for a buffer overfill.
    \item \textbf{agent training:}
    In the training phase the agent interacts with the training environment that was previously defined in the procedure. The agent uses \ac{rl} algorithms to learn to make decisions that optimize its received reward.
    \item \textbf{test environment:}
    A test environment may be developed to simulate the production environment and test the agent under controlled conditions, thereby reducing the risk of production downtime. The development and optimization of the agent is conducted in cooperation with experienced production planners of the manufacturer.
    Conducting a pilot project in parallel with the existing systems allows for the evaluation of the agent's decisions and its impact on production processes.
    \item \textbf{deployment and scaling:}
    Once the agent has been trained it can be deployed in the production system. This process is conducted in a systematic and sequential manner, starting with less critical processes in order to minimize potential risks. A feedback loop is established for the continuous evaluation of the agents' performance and adaptation to the changing environment, based on real production data.
    \item \textbf{employee training and change management:}
    Staff is trained on the use of the new system, to check the proposed schedules, and override the decisions made by the agent. The implementation of change management strategies ensures the effective utilization of the new system by the employees and strengthens their acceptance of these new technologies.
    \item \textbf{maintenance and continuous improvement:}
    Regular reviews and adjustments of the agent ensure an optimal performance and adaptation to the constantly changing production conditions.
\end{enumerate}

\section{Conclusion and Outlook}\label{cha:Conclusion}

This paper presents a concept for \ac{drl}-supported production planning that can be adapted for optimized job shop scheduling. In Chapter \ref{cha:Method}, the elements of the \ac{drl} model are presented, including factors that aim to bridge the gap between real-world production environments and production models. The consideration of fluctuating job volumes throughout the production process ensures more precise estimations of the required space conditions within the buffer zones across the shop floor. This prevents overfilled buffer areas, which clog the production flow as well as the accumulation of unnecessary materials and capital lockup. The introduction of the quantity factor $\delta$ determines the processing times of each operation, depending on its batch size, while transport times are added to represent the production duration more accurately. Jobs are specified with a deadline to simulate a delivery date, while machines are described with machine setups, that represent their current state in a more realistic manner. Furthermore, changes in the setup are also taken into account for the processing duration.
The introduction of these framework conditions enables the generic \ac{jssp} model to be extended in a manner that more accurately reflects the high complexity of a real-world production environment. This extended approach allows the construction of a training environment, in which a \ac{rl} agent can learn to optimally set up machines, assign jobs to machines, and move jobs between machines and buffers in accordance with specific optimization goals.

This approach is applied to the industrial furniture production in Chapter \ref{cha:Integration}, tackling industry-specific challenges including the translation from a real-world furniture article to a job in the model. Two implementation concepts are presented for the integration of an \ac{rl} agent into production planning systems: episodic and continuous planning systems. The production planning process for an episodic planning approach is illustrated using an example \ac{jssp}, highlighting the functionality of a trained agent at different time steps in the planning process.
While episodic planning can be integrated as a low-tech, standalone solution, a continuous planning agent is fully integrated into the existing production systems. This enables real-time scheduling and allows for the prompt reaction to machine breakdowns as soon as the error message appears in the \ac{mes}. In the event of predicted material shortages, as indicated by the data from the \ac{erp} system, the production of specific articles may be postponed in accordance with the anticipated shortage. However, the interface communication between the agent and the production system, as well as the implementation of the agent into the system, require a complex, customized solution.

A challenge remains in precisely defining a reward function that considers the proposed elements and the industry specific optimization goals. It remains unclear how to best prevent buffer overfill, particularly given that buffer levels are only checked at the time steps in episodic planning, rather than in between time steps. It remains to be investigated whether an overfilled buffer causes a deadlock of the system when a job cannot be moved to the next buffer because it is full. The consideration of the batch size for the processing duration with the introduced quantity factor $\delta$ can be extended to the calculation of the required transportation times. Larger batch sizes may include more pallets to be moved from one place to another, which would result in longer transportation times. A further challenge arises when the production system is not organised as a generic job shop but as a dynamic job shop: When the same furniture components can be produced by different machines on several routes and thus alternative routes through production are possible. This significantly increases the level of complexity, as the number of possible combinations grows at an even faster rate than in a generic job shop.
The following step of this process is the realisation of the concept described above in order to examine, how an agent would deal with the increased level of complexity and increasing problem sizes. It is also important to note that the scheduling agents described above have been designed to support human production planners, rather than creating artificial copies of them.\cite{Klarmann.2021}

\printacronyms[name=Abbreviations]



\begin{thebibliography}{10}

\bibitem{Aydin.2000}
M.~Aydin and E.~{\"O}ztemel.
\newblock Dynamic job-shop scheduling using reinforcement learning agents.
\newblock {\em Robotics and Autonomous Systems}, 33(2-3), 2000.

\bibitem{Bazewicz.2001}
J.~B{\l}a{\.z}ewicz, K.~H. Ecker, E.~Pesch, G.~Schmidt, and J.~W{\k{e}}glarz.
\newblock {\em Scheduling Computer and Manufacturing Processes}.
\newblock {Springer Berlin Heidelberg}, Berlin, Heidelberg, 2001.

\bibitem{Deale.1994}
M.~Deale, M.~Yvanovich, D.~Schnitzuius, D.~Kautz, M.~Carpenter, M.~Zweben, G.~Davis, and B.~Daun.
\newblock The space shuttle ground processing scheduling system.
\newblock {\em Intelligent Scheduling}, pages 423--449, 1994.

\bibitem{Gabel.2012}
T.~Gabel and M.~Riedmiller.
\newblock Distributed policy search reinforcement learning for job-shop scheduling tasks.
\newblock {\em International Journal of Production Research}, 50(1):41--61, 2012.

\bibitem{Garey.1976}
M.~R. Garey, D.~S. Johnson, and R.~Sethi.
\newblock The complexity of flowshop and jobshop scheduling.
\newblock {\em Mathematics of Operations Research}, 1(2):117--129, 1976.

\bibitem{Han.2020}
B.-A. Han and J.-J. Yang.
\newblock Research on adaptive job shop scheduling problems based on dueling double dqn.
\newblock {\em IEEE Access}, 8:186474--186495, 2020.

\bibitem{Klarmann.2021}
N.~Klarmann.
\newblock Artificial intelligence narratives: An objective perspective on current developments.

\bibitem{Lillicrap.2015}
T.~P. Lillicrap, J.~J. Hunt, A.~Pritzel, N.~Heess, T.~Erez, Y.~Tassa, D.~Silver, and D.~Wierstra.
\newblock Continuous control with deep reinforcement learning.

\bibitem{Liu.2020}
C.-L. Liu, C.-C. Chang, and C.-J. Tseng.
\newblock Actor-critic deep reinforcement learning for solving job shop scheduling problems.
\newblock {\em IEEE Access}, 8:71752--71762, 2020.

\bibitem{OpenAI.}
OpenAI.
\newblock Openai five.

\bibitem{Oren.442021}
J.~Oren, C.~Ross, M.~Lefarov, F.~Richter, A.~Taitler, Z.~Feldman, C.~Daniel, and D.~{Di Castro}.
\newblock Solo: Search online, learn offline for combinatorial optimization problems.

\bibitem{Panzer.2022}
M.~Panzer and B.~Bender.
\newblock Deep reinforcement learning in production systems: a systematic literature review.
\newblock {\em International Journal of Production Research}, 60(13):4316--4341, 2022.

\bibitem{Park.2021}
J.~Park, J.~Chun, S.~H. Kim, Y.~Kim, and J.~Park.
\newblock Learning to schedule job-shop problems: representation and policy learning using graph neural network and reinforcement learning.
\newblock {\em International Journal of Production Research}, 59(11):3360--3377, 2021.

\bibitem{Pezzella.2008}
F.~Pezzella, G.~Morganti, and G.~Ciaschetti.
\newblock A genetic algorithm for the flexible job-shop scheduling problem.
\newblock {\em Computers {\&} Operations Research}, 35(10):3202--3212, 2008.

\bibitem{Pinedo.2012}
M.~L. Pinedo.
\newblock {\em Scheduling}.
\newblock {Springer US}, Boston, MA, 2012.

\bibitem{Rinciog.2022}
A.~Rinciog and A.~Meyer.
\newblock Towards standardising reinforcement learning approaches for production scheduling problems.
\newblock {\em Procedia CIRP}, 107:1112--1119, 2022.

\bibitem{SerranoRuiz.2024}
J.~C. Serrano-Ruiz, J.~Mula, and R.~Poler.
\newblock Job shop smart manufacturing scheduling by deep reinforcement learning.
\newblock {\em Journal of Industrial Information Integration}, 38:100582, 2024.

\bibitem{Shahrabi.2017}
J.~Shahrabi, M.~A. Adibi, and M.~Mahootchi.
\newblock A reinforcement learning approach to parameter estimation in dynamic job shop scheduling.
\newblock {\em Computers {\&} Industrial Engineering}, 110:75--82, 2017.

\bibitem{Silver.2016}
D.~Silver, A.~Huang, C.~J. Maddison, A.~Guez, L.~Sifre, G.~{van den Driessche}, J.~Schrittwieser, I.~Antonoglou, V.~Panneershelvam, M.~Lanctot, S.~Dieleman, D.~Grewe, J.~Nham, N.~Kalchbrenner, I.~Sutskever, T.~Lillicrap, M.~Leach, K.~Kavukcuoglu, T.~Graepel, and D.~Hassabis.
\newblock Mastering the game of go with deep neural networks and tree search.
\newblock {\em Nature}, 529(7587):484--489, 2016.

\bibitem{Sutton.2018}
R.~S. Sutton and A.~Barto.
\newblock {\em Reinforcement learning: An introduction}.
\newblock Adaptive computation and machine learning. {The MIT Press}, Cambridge, Massachusetts and London, England, second edition edition, 2018.

\bibitem{Taillard.1994}
{\'E}.~D. Taillard.
\newblock Parallel taboo search techniques for the job shop scheduling problem.
\newblock {\em ORSA Journal on Computing}, 6(2):108--117, 1994.

\bibitem{Tassel.482021}
P.~Tassel, M.~Gebser, and K.~Schekotihin.
\newblock A reinforcement learning environment for job-shop scheduling.

\bibitem{vanLaarhoven.1992}
P.~J.~M. {van Laarhoven}, E.~H.~L. Aarts, and J.~K. Lenstra.
\newblock Job shop scheduling by simulated annealing.
\newblock {\em Operations Research}, 40(1):113--125, 1992.

\bibitem{Vivekanandan.2023}
D.~Vivekanandan, S.~Wirth, P.~Karlbauer, and N.~Klarmann.
\newblock A reinforcement learning approach for scheduling problems with improved generalization through order swapping.
\newblock {\em Machine Learning and Knowledge Extraction}, 5(2):418--430, 2023.

\bibitem{Wang.2021}
L.~Wang, X.~Hu, Y.~Wang, S.~Xu, S.~Ma, K.~Yang, Z.~Liu, and W.~Wang.
\newblock Dynamic job-shop scheduling in smart manufacturing using deep reinforcement learning.
\newblock {\em Computer Networks}, 190:107969, 2021.

\bibitem{Waschneck.2018}
B.~Waschneck, A.~Reichstaller, L.~Belzner, T.~Altenm{\"u}ller, T.~Bauernhansl, A.~Knapp, and A.~Kyek.
\newblock Optimization of global production scheduling with deep reinforcement learning.
\newblock {\em Procedia CIRP}, 72:1264--1269, 2018.

\bibitem{Zhang.2020}
C.~Zhang, W.~Song, Z.~Cao, J.~Zhang, P.~S. Tan, and X.~Chi.
\newblock Learning to dispatch for job shop scheduling via deep reinforcement learning.
\newblock {\em Advances in neural information processing systems}, 33:1621--1632, 2020.

\bibitem{Zhang.1995}
W.~Zhang and T.~G. Dietterich.
\newblock A reinforcement learning approach to job-shop scheduling.
\newblock In {\em Ijcai}, volume~95, pages 1114--1120, 1995.

\bibitem{Zhao.2021}
Y.~Zhao and H.~Zhang.
\newblock Application of machine learning and rule scheduling in a job-shop production control system.
\newblock {\em International Journal of Simulation Modelling}, 20(2):410--421, 2021.

\end{thebibliography}
\end{document}